\documentclass[10pt,twocolumn,letterpaper]{article}

\usepackage{cvpr}
\usepackage{times}
\usepackage{epsfig}
\usepackage{graphicx}
\usepackage{amsmath}
\usepackage{amssymb}

\usepackage[breaklinks=true,bookmarks=false]{hyperref}

\cvprfinalcopy 

\def\cvprPaperID{****} 

\begin{document}

\title{Towards Good Practices for Deep 3D Hand Pose Estimation}

\author{Hengkai Guo\\
Tsinghua University\\
Beijing, China\\
{\tt\small guohengkaighk@gmail.com}
\and
Guijin Wang\\
Tsinghua University\\
Beijing, China\\
{\tt\small wangguijin@tsinghua.edu.cn}
\and
Xinghao Chen\\
Tsinghua University\\
Beijing, China\\
{\tt\small chenxinghao1010@163.com}
\and
Cairong Zhang\\
Tsinghua University\\
Beijing, China\\
{\tt\small zcr13@mails.tsinghua.edu.cn}
}

\maketitle

\begin{abstract}
3D hand pose estimation from single depth image is an important and challenging problem for human-computer interaction. Recently deep convolutional networks (ConvNet) with sophisticated design have been employed to address it, but the improvement over traditional random forest based methods is not so apparent. To exploit the good practice and promote the performance for hand pose estimation, we propose a tree-structured Region Ensemble Network (REN) for directly 3D coordinate regression. It first partitions the last convolution outputs of ConvNet into several grid regions. The results from separate fully-connected (FC) regressors on each regions are then integrated by another FC layer to perform the estimation. By exploitation of several training strategies including data augmentation and smooth $L_1$ loss, proposed REN can significantly improve the performance of ConvNet to localize hand joints. The experimental results demonstrate that our approach achieves the best performance among state-of-the-art algorithms on three public hand pose datasets. We also experiment our methods on fingertip detection and human pose datasets and obtain state-of-the-art accuracy.
\end{abstract}

\section{Introduction}
3D hand pose estimation from depth imaging has drawn lots of attention from researchers \cite{sinha2016deephand} \cite{ye2016spatial} \cite{wan2016hand} due to its important role in applications of augmented reality (AR) and human-computer interface (HCI) \cite{zhou2016novel}. It aims to predict the 3D accurate positions for hand joints \cite{supancic2015depth} with monocular depth images, which is critical for gesture recognition \cite{chen2016static}. Though has been studied for several years \cite{supancic2015depth}, it is still challenging owing to high joint flexibility, large view variance, poor depth quality, severe self occlusion, and similar part confusion.

Recently, deep convolutional networks (ConvNets) have exhibited state-of-the-art performance across several computer vision tasks such as object classification \cite{krizhevsky2012imagenet}, object detection \cite{girshick2016region}, and image segmentation \cite{chen2016deeplab}. ConvNets have also been employed to solve the problem of hand pose estimation, often with complicated structure design such as multi-branch inputs \cite{tompson2014real}\cite{oberweger2015hands} and multi-model regression \cite{oberweger2015hands} \cite{oberwegertraining} \cite{gerobust} \cite{zhang2016learning}. Thanks to the great modeling capacity and end-to-end feature learning, deep ConvNets have achieved competitive accuracy for hand pose estimation. However, ConvNets remain unable to obtain significant advantage over traditional random forest based methods \cite{sun2015cascaded} \cite{wan2016hand}, which may result from the relatively shallow ConvNet structure (often 3 - 5 convolution layers \cite{tompson2014real} \cite{oberwegertraining} \cite{zhang2016learning}) and high risk of overfitting with relative small datasets compared to image classification.

\begin{figure}[htb]
\centering
{\includegraphics[width=0.48\textwidth]{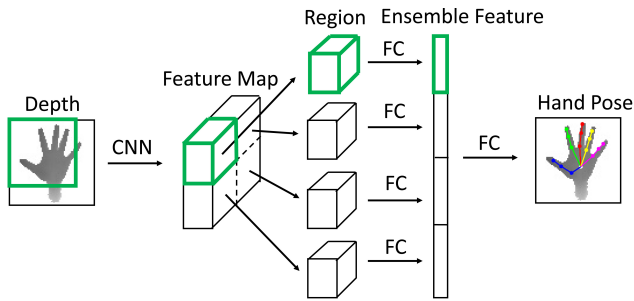}}
\caption{Region ensemble network (REN) with four regions: First deep ConvNet is used to extract features of depth image. The feature maps from ConvNet are then divided into regions. Each region is finally fed into fully-connected (FC) layers and then fused to predict the hand pose. The green rectangles represent the receptive field of the top-left region on the feature maps.}
\label{fig_overview}
\end{figure}

In this paper, we explore multiple good practices with hand pose estimation in single depth images. Most importantly, inspired by model ensemble and multi-view voting \cite{krizhevsky2012imagenet}, we present a single deep ConvNet architecture named \emph{Region Ensemble Net (REN)}\footnote{Codes and models are available at \url{https://github.com/guohengkai/region-ensemble-network}} (Fig.\ref{fig_overview}) to directly regress the 3D hand joint coordinates with end-to-end optimization and inference. We implement it by training individual fully-connected (FC) layers on multiple feature regions and combining them as ensembles. In addition, we adopt several approaches to enhance the performance including residual connection \cite{he2015deep}, data augmentation and smooth $L_1$ loss \cite{girshick2015fast}. As shown in our experiments, REN significantly promotes the performance of our ConvNet, which outperforms state-of-the-art methods on three challenging hand pose benchmarks \cite{tompson2014real} \cite{tang2014latent} \cite{sun2015cascaded}. Evaluated on fingertip \cite{tompson2014real} and human pose benchmarks \cite{haque2016towards}, our REN also achieves the best accuracy.

This paper builds on our preliminary publication \cite{guo2017region}. Compared with it, this paper describes more techinical details and discusses several important factors for good practice, leading to slightly better results than \cite{guo2017region} with different region settings. In addition, we add results for one extra hand pose dataset \cite{sun2015cascaded}, and further evaluate our REN for fingertip detection and human pose estimation with state-of-the-art performance.

\section{Related work}
We briefly review relevant hand pose estimation methods with ConvNets for depth imaging, and examine methodologies related to the proposed algorithm, including ensemble methods and multi-view testing for ConvNets. Finally we also introduce works using ConvNets for RGB-D fingertip detection and human pose estimation, which will be compared in our experiments.

\subsection{RGB-D Hand pose estimation with ConvNets}
Recently deep ConvNets have been applied on hand pose estimation for depth imaging. Tompson et al. \cite{tompson2014real} first use ConvNets to produce 2D heat maps with multi-scale inputs and infer the 3D hand pose with inverse kinematics. Oberweger et al. \cite{oberweger2015hands} directly regress the 3D joint locations with multi-scale and multi-stage ConvNets using a linear layer as pose prior. In \cite{oberwegertraining}, a feedback loop is employed to iteratively correct the mistakes of inference, in which three ConvNets are used for pose initialization, image synthesis and pose updating. Ge et al. \cite{gerobust} employ three ConvNets from orthogonal views to separately regress 2D heat maps for each views with depth projections and fuse them to produce 3D hand pose. In \cite{zhou2016model}, physical joint constraints are incorporated into a forward kinematics based layer in ConvNet. Similarly, Zhang et al. \cite{zhang2016learning} embed skeletal manifold into ConvNets and train the model end-to-end to render sequential prediction.

\subsection{Multi-model ensemble methods for ConvNets}
Traditional ensemble learning means that training multiple individual models and combining their outputs via averaging or weighted fusions, which is widely adopted in recognition competitions \cite{krizhevsky2012imagenet}. In addition to bagging \cite{krizhevsky2012imagenet} \cite{sun2013deep}, boosting is also introduced for people counting \cite{walach2016learning}. However, using multiple ConvNets for both training and testing requires huge cost of memory and time, which is not practical for applications.

\subsection{Multi-branch ensemble methods for ConvNets}
Single ConvNet with the fusion of multiple branches can also be regarded as a generalized type of ensemble. One popular strategy is to fuse different scaling inputs \cite{tompson2014real} \cite{oberweger2015hands} or different image cues \cite{guo2016two} \cite{li2016deeptrack} \cite{chen2016accurate} with multi-input branches. Another approach is to employ multi-output branches with shared convolutional feature extractor, either training with different samples \cite{li2016convolutional} or learning to predict different categories \cite{ahmed2016network}. Compared with multi-input ensemble, multi-output methods cost less time because inference of FC layers is much faster than that of convolutional layers. Our method also falls into such category, but we apply ensemble on feature regions instead of inputs.

\subsection{Multi-view testing for ConvNets}
Multi-view testing is widely adopted to improve accuracy for object classification \cite{krizhevsky2012imagenet} \cite{sermanet2013overfeat} \cite{he2014spatial}. In \cite{krizhevsky2012imagenet}, predictions from 10-crop (four corners and one center with horizontal flip) are averaged on single ConvNet. In \cite{sermanet2013overfeat} \cite{he2014spatial}, fully-convolutional networks are employed in testing with multi-scale and multi-view inputs. Then spatially average pooling is applied on the class score map to obtain the final scores. To best of our knowledge, such strategy has not been applied on 3D pose regression yet.

\subsection{RGB-D fingertip detection and human pose estimation for ConvNets}
Fingertips play an important role in human-computer interaction among the hand joints. Wetzler et al. \cite{wetzler2015rule} employ ConvNet for in-plane derotation of hand depth image and then use random forests or ConvNets for fingertip coordinate regression. Guo et al. \cite{guo2016two} introduce a two-stream ConvNet to detect the 3D fingertips, which makes use of both depth information and edge information with slow fusion strategy.

Human pose estimation is also important for HCI applications such as action recognition \cite{zhang2016rgb} \cite{chen2016novel}. Though ConvNets are widely used in human pose estimation for RGB images \cite{carreira2016human} \cite{bulat2016human}, there are limited number of works using ConvNets for depth imaging \cite{shi2015high} \cite{wang2013depth} due to relatively small size of training datasets. Haque et al. \cite{haque2016towards} introduce a viewpoint invariant model using ConvNets and recurrent networks (RNNs) for human pose estimation. Local regions from depth images are transformed into a learned feature space via ConvNets and then RNNs are leveraged to predict the offsets of pose sequentially with multi-task setting.

\section{Region Ensemble Network}
As in Fig.\ref{fig_overview}, REN starts with a ConvNet for feature extraction. Then the features are divided into multiple grid regions. Each region is fed into FC layers and learnt to fuse for hand pose prediction. In this section we introduce the basic network architecture, region ensemble structure and implementation details.

\subsection{Network architecture with residual connection}
\label{residual}
\begin{figure}[htb]
\centering
{\includegraphics[width=0.48\textwidth]{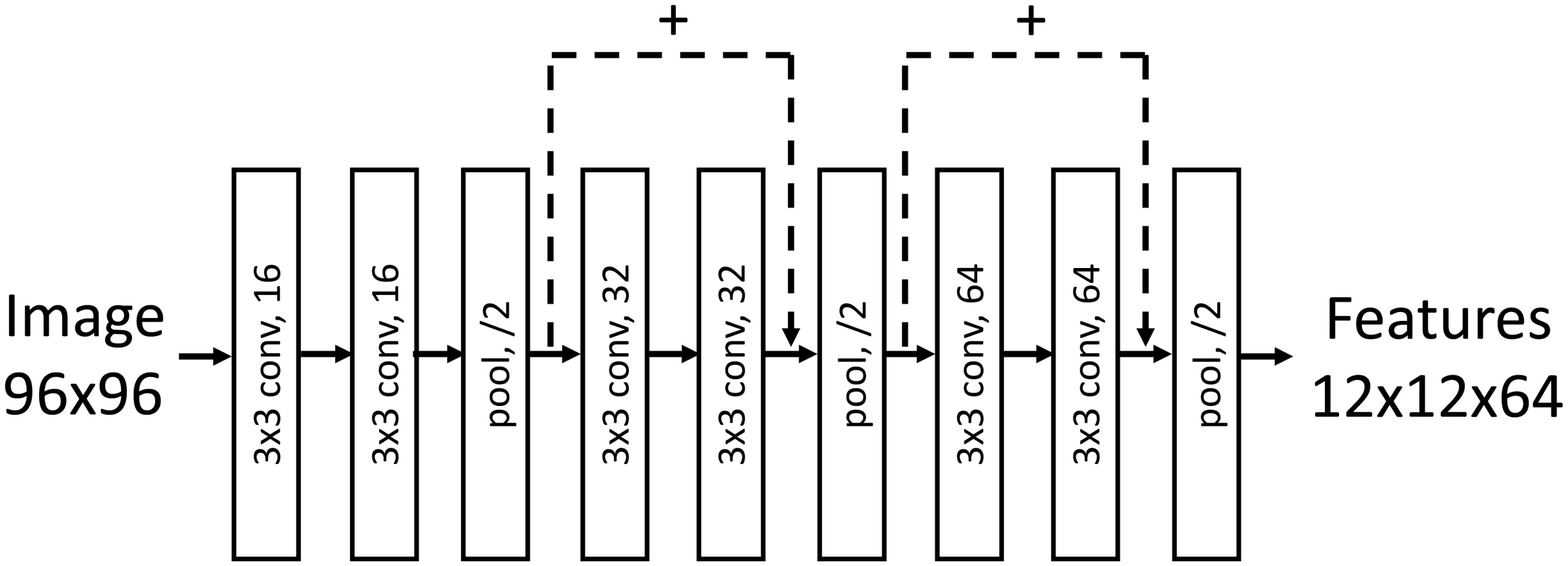}}
\caption{Structure of basic ConvNet for feature extraction. The ConvNet consists of six convolutional layers and three pooling layers. The dotted arrows represent residual connections with dimension increase \cite{he2015deep}. The non-linear activation layers following each convolutional layers are not showed in the figure.}
\label{fig_baseline}
\end{figure}

The architecture of our ConvNet for feature extraction consists of six convolutional layers with $3\times3$ kernels (Fig.\ref{fig_baseline}) and three pooling layers with $2\times2$ kernels. Each convolutional layer is followed by a Rectified Linear Unit (ReLU) activation. The ConvNet accepts inputs of a $96\times96$ depth image and outputs the feature maps with dimension of $12\times12\times64$. To improve the learning ability, two residual connections are adopted between pooling layers with $1\times1$ convolution filters for dimension increase as in \cite{he2015deep}. So there are totally eight convolutional layers in our model, which is deeper than ConvNets in \cite{zhang2016learning} with five layers.

For regression, we use two 2048 dimension FC layers with dropout rate \cite{srivastava2014dropout} of 0.5 for each regressor to avoid overfitting. The output of regressor is a $3\times J$ vector representing the 3D world coordinates for hand joints, where $J$ is the number of joints.

\subsection{Region ensemble structure}
\label{ensemble}
Multi-view testing averages predictions from different crops of original input image, which reduces the variance for image classification \cite{krizhevsky2012imagenet}. Because image classification is invariant to translation and cropping, multi-view testing is easy to apply by directly cropping on the input image. When it comes to pose regression, each cropped parts will correspond to different hand pose configurations. So we should adapt the 3D coordinates of hand pose to the cropped view. Meanwhile, using multiple inputs to feed the ConvNet one-by-one is time-consuming.

Because each activation in the convolutional feature maps is contributed by a receptive field in the input image domain, we can project the multi-view inputs onto the regions of the feature maps. By using separate regions as features, we can train separate regressors instead of single regressor. So multi-view voting could be extended to regression task by utilizing each regions to separately predict the whole hand pose and then combining the results.

Based on this inspiration, we define a tree-structured network consisting of a single ConvNet trunk and several regression branches as shown in Fig.\ref{fig_overview}. We first divide the feature maps of ConvNet into several regions. For each region, we feed it into the FC layers respectively as branches. There are several ways to combine different branches. A simple strategy is bagging, which averages all outputs of branches using average pooling. In order to boost the predictions from all the regions, we employ region ensemble strategy instead of bagging: features from the last FC layers of all regions are concatenated and used to infer the coordinates with an extra regression layer. The whole network can be trained end-to-end by minimizing the regression loss.

\begin{figure}[htb]
\centering
{\includegraphics[width=0.48\textwidth]{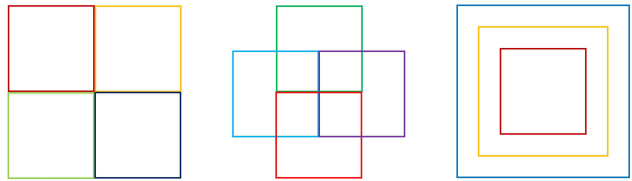}}
\caption{Different region setting for feature maps: four conners \cite{guo2017region} (left), four centers in each edges (middle), and multi-scale regions with the same center (right). Proposed REN adopts nine regions with size of $6\times6$ including the center of feature maps and all the eight regions in left and middle figures. }
\label{fig_region}
\end{figure}

\begin{figure}[htb]
\centering
{\includegraphics[width=0.48\textwidth]{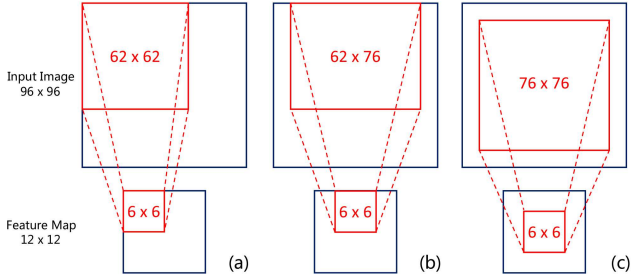}}
\caption{Receptive fields for different region positions: (a) $62\times62$ for conners, (b) $62\times76$ or $76\times62$ for centers in each edges, and (c) $76\times76$ for the center of feature maps.}
\label{fig_receptive}
\end{figure}

For region setting, we use nine regions with size of $6\times6$ located at four conners (left part in Fig. \ref{fig_region}, which is also the whole setting in \cite{guo2017region}), four centers near the edges (middle part in Fig. \ref{fig_region}) and the center of the feature maps. The receptive fields of different regions within the $96\times96$ image bounding are shown in Fig. \ref{fig_receptive}, which is similar to the corner and center crop in \cite{krizhevsky2012imagenet}. We will discuss the effect of different region settings on accuracy in Section \ref{setting}.

There are three main differences between proposed methods and multi-view voting: 1) To our knowledge, all multi-view testing methods before are designed for image classification while our region ensemble can be applied on both classification and regression. By applying fusion FC layer in REN, different views of inputs are trained to simultaneously predict the same pose. 2) We adopt end-to-end training for region ensemble instead of testing only, making the ConvNet adjust the contributions from each views. 3) We replace the average pooling with one FC layer on concatenated features to learn the fusion parameters, which increases the learning ability of the network. We will perform the comparison in Section \ref{basic}.

\subsection{Implementation details}
\label{detail}
We implement our REN with Caffe \cite{jia2014caffe} written in C++. We use stochastic gradient descent (SGD) with a mini-batch size of 128. The learning rate starts from 0.005 and is divided by 10 every 20 epochs, and the model is trained for total 80 epochs. In the meanwhile, we use a weight decay of 0.0005 and a momentum of 0.9. Our model is trained from scratch with random initialization \cite{he2015delving}. Moreover, there are three important strategies for training: patch cropping, data augmentation, and smooth $L_1$ loss. The details are described below. And we will show the incremental contributions of these strategies in later section.

\noindent\textbf{Patch cropping}\hspace{2mm} For ConvNet inputs, we extract a cube with fixed size of 150mm from the depth image centered in the hand region. Then the cube is resized into a $96\times96$ patch of depth values normalized to $[-1, 1]$ as input for ConvNet. The 3D coordinates are also normalized to $[-1, 1]$ according to the cube. To compute the center, we first segment the foreground with fixed thresholds and calculate the centroid of foreground.

\noindent\textbf{Data augmentation}\hspace{2mm} We apply data augmentation during training, including translation within $[-10, 10]$ pixel, scaling within $[0.9, 1.1]$ and rotation within $[-180, 180]$ degree. Random augmentation effectively increases the size of training dataset, so it can improve the generalization performance.

\noindent\textbf{Smooth $L_1$ loss}\hspace{2mm} To deal with noisy annotations, we adopt similar smooth $L_1$ loss in \cite{girshick2015fast}:
\begin{displaymath}
\textrm{smooth}_{L_1}(x)=\left\{ \begin{array}{ll}
	0.5x^2 & \textrm{if} |x| < 0.01 \\
	0.01(|x| - 0.005) & \textrm{otherwise}
\end{array} \right.
\end{displaymath}
Because it is less sensitive to outliers than the $L_2$ loss, it can benefit the training of ConvNet.

\section{Experiments}
In this section, we first introduce the evaluation datasets and metrics for our experiments. Then we compare our REN with several state-of-the-art methods on public hand pose datasets. Next we explore several good practices of training ConvNets for hand pose estimation, discuss different region settings and also compare with traditional ensembles and multi-view testing. Finally we apply our REN on fingertip detection and human pose estimation for public benchmarks. 

\subsection{Experiment setup}
\subsubsection{Datasets}
We conduct our experiments on four publicly RGB-D datasets: ICVL hand pose dataset \cite{tang2014latent}, NYU hand pose dataset \cite{tompson2014real}, MSRA hand pose dataset \cite{sun2015cascaded}, and ITOP human pose dataset \cite{haque2016towards}. For self-comparison, ICVL dataset is mainly used. More details for datasets are as follows:

\noindent\textbf{ICVL dataset}\hspace{2mm} The training set of ICVL dataset contains 300K images with different rotations, and the testing set contains 1.6K images. All the depth images are captured by Intel RealSense. Totally 16 hand joints are initialized by the output of camera and manually refined.

\noindent\textbf{NYU dataset}\hspace{2mm} The NYU dataset has 72K images for training and 8K for testing with 36 3D annotated joints, collected from Microsoft Kinect camera. Following \cite{tompson2014real}, 14 hand joints with front-view image are used in experiments. And this dataset is also used to evaluate fingertip detection on the 5 fingertip joints in \cite{wetzler2015rule} \cite{guo2016two}.

\noindent\textbf{MSRA dataset}\hspace{2mm} The MSRA dataset contains 9 subjects with 17 gestures for each subject. 76K depth images with 21 annotated joints are collected with Intel's Creative Interactive Camera. For evaluation, each subject is alternatively used as testing data when other 8 subjects are used for training. This is repeated 9 times and the average metrics are reported.

\noindent\textbf{ITOP dataset}\hspace{2mm} The ITOP dataset consists 18K training images and 5K testing images for front view and top view acquired by two Kinect cameras. Each depth image is labelled with fifteen 3D joint locations of human body.

\subsubsection{Evaluation metrics}
We employ different metrics for hand pose estimation and human pose estimation following the literatures \cite{tang2014latent} \cite{tompson2014real} \cite{haque2016towards}. For hand pose, the performance is evaluated by two metrics: 1) \textbf{average 3D distance error} is computed as the average Euclidean distance for each joint (in millimeters). 2) \textbf{percentage of success frames} is defined as the rate of frames in which all Euclidean errors of joints are below a variant threshold \cite{oberweger2015hands}. In addition, mean precision (mP) with a threshold of 15mm as defined in \cite{wetzler2015rule} is calculated for fingertip detection.

For human pose, we compute the \textbf{mean average precision (mAP)} \cite{haque2016towards}, which is defined as the average detected rate for all human body joints. A joint is counted as detected when the Euclidean distance between predicted position and ground truth is below 10cm.

\subsection{Comparison with the state of the art}
\begin{figure*}[htb]
\centering
\begin{minipage}[b]{0.49\textwidth}
  \centering
  \centerline{\includegraphics[width=0.95\textwidth]{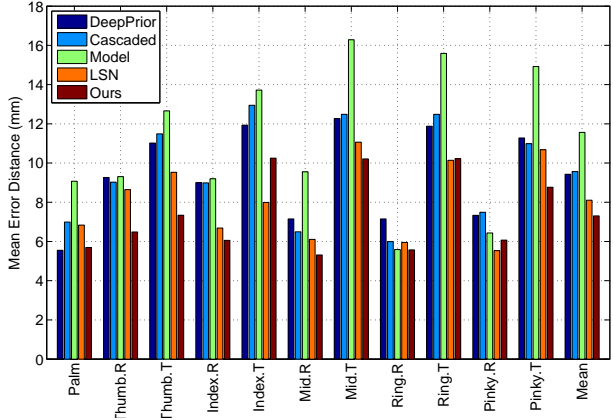}}
\end{minipage}
\begin{minipage}[b]{0.49\textwidth}
  \centering
  \centerline{\includegraphics[width=0.95\textwidth]{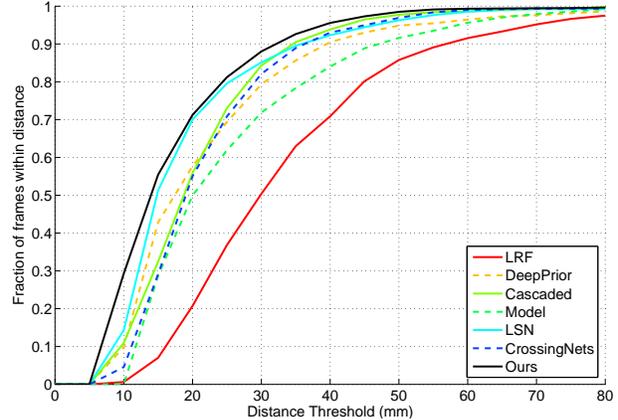}}
\end{minipage}
\caption{Comparison with state-of-the-arts on ICVL \cite{tang2014latent} dataset: distance error (left) and percentage of success frames (right).}
\label{fig_sota_icvl}
\end{figure*}

\begin{figure*}[htb]
\begin{minipage}[b]{0.49\textwidth}
  \centering
  \centerline{\includegraphics[width=0.95\textwidth]{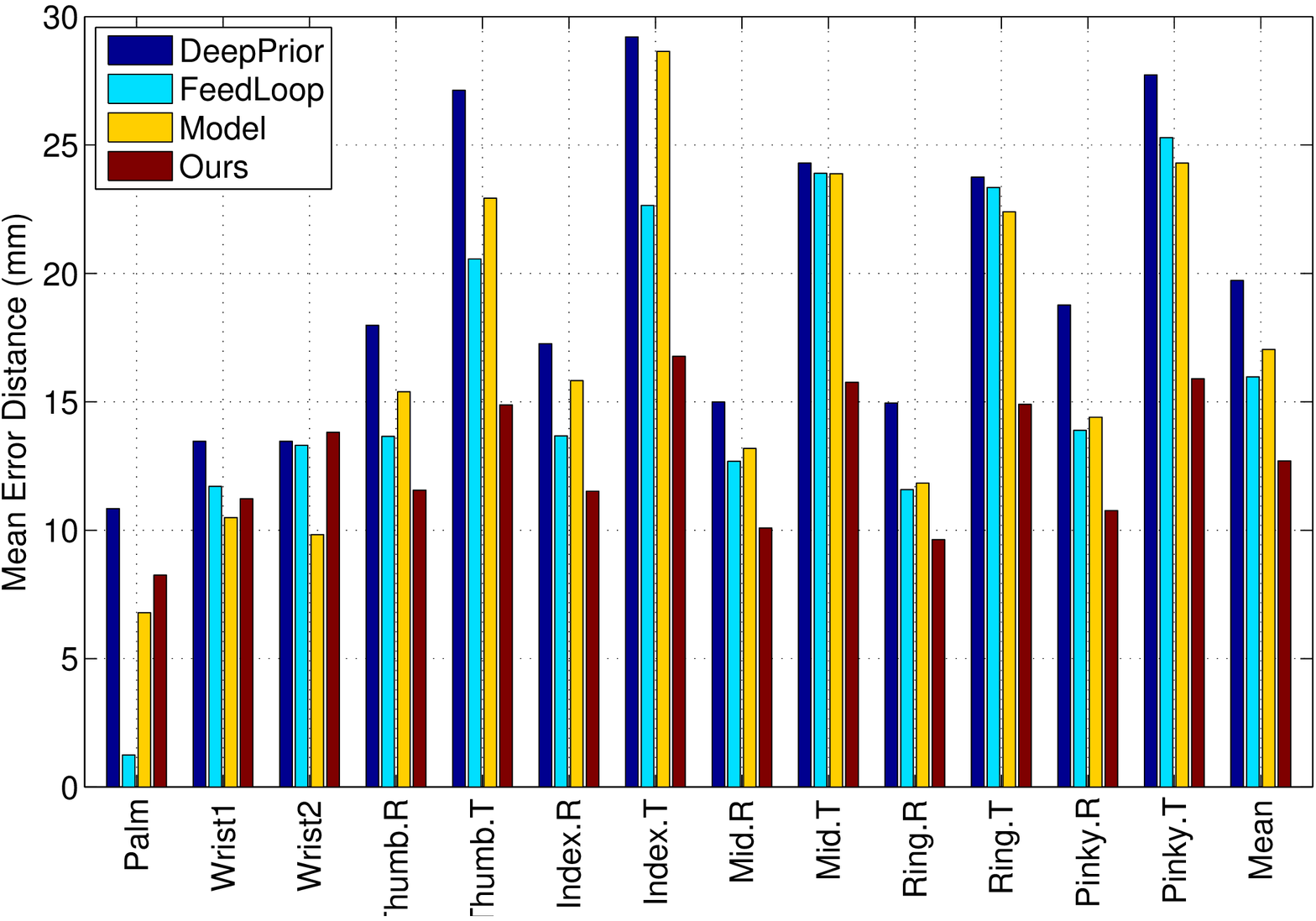}}
\end{minipage}
\begin{minipage}[b]{0.49\textwidth}
  \centering
  \centerline{\includegraphics[width=0.95\textwidth]{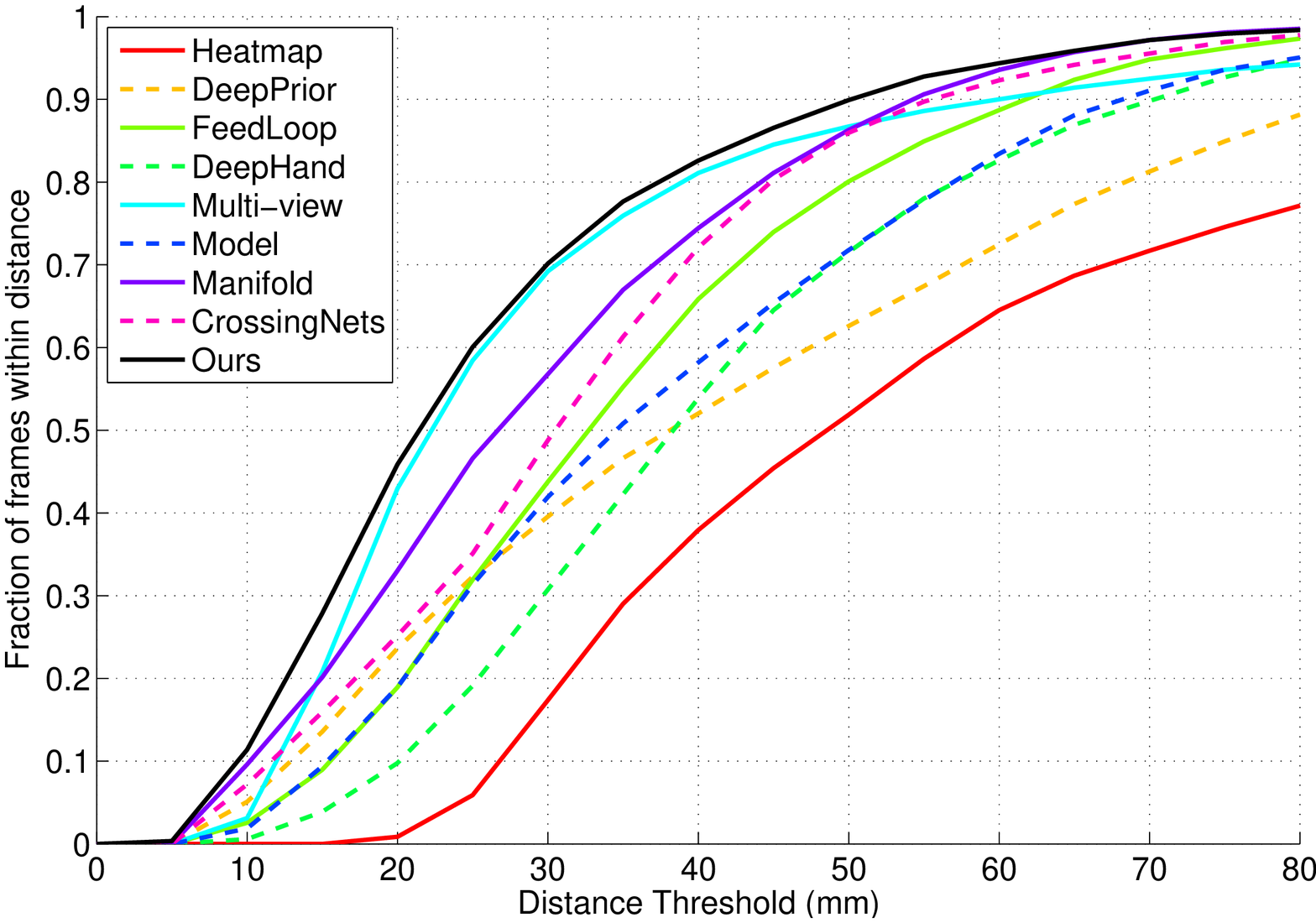}}
\end{minipage}
\caption{Comparison with state-of-the-arts on NYU \cite{tompson2014real} datasets: distance error (left) and percentage of success frames (right).}
\label{fig_sota_nyu}
\end{figure*}

\begin{figure*}[htb]
\begin{minipage}[b]{0.49\textwidth}
  \centering
  \centerline{\includegraphics[width=0.95\textwidth]{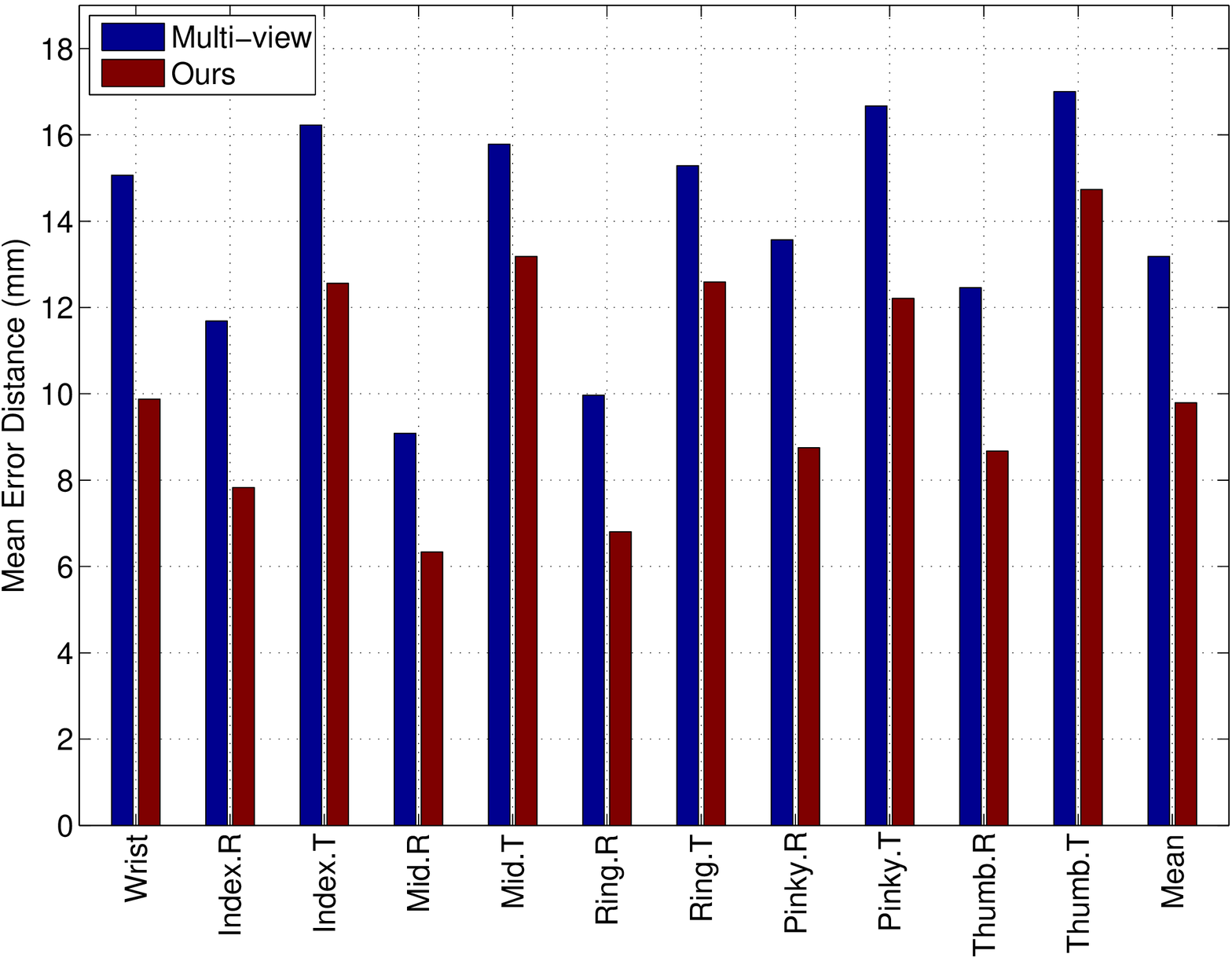}}
\end{minipage}
\begin{minipage}[b]{0.49\textwidth}
  \centering
  \centerline{\includegraphics[width=0.95\textwidth]{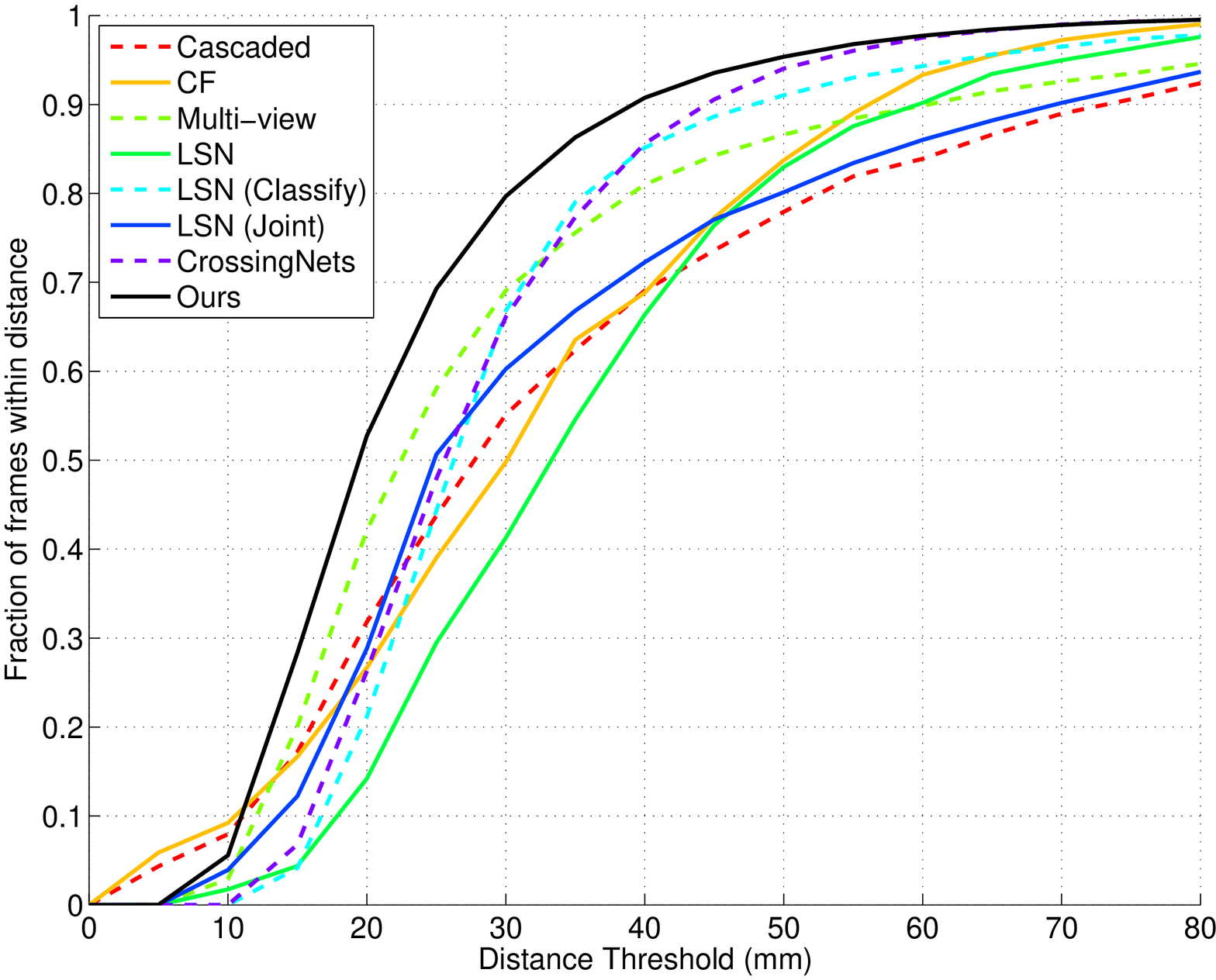}}
\end{minipage}
\caption{Comparison with state-of-the-arts on MSRA \cite{sun2015cascaded} datasets: distance error (left) and percentage of success frames (right).}
\label{fig_sota_msra}
\end{figure*}

We compare our methods against several state-of-the-art approaches on ICVL dataset \cite{tang2014latent} \cite{oberweger2015hands} \cite{sun2015cascaded} \cite{zhou2016model} \cite{wan2016hand} \cite{wan2017cross}, NYU dataset \cite{tompson2014real} \cite{oberweger2015hands} \cite{oberwegertraining} \cite{sinha2016deephand} \cite{gerobust} \cite{zhou2016model} \cite{zhang2016learning} \cite{wan2017cross}, and MSRA dataset \cite{sun2015cascaded} \cite{choi2015collaborative} \cite{gerobust} \cite{wan2016hand} \cite{wan2017cross}. Overall, Fig.\ref{fig_sota_icvl} - \ref{fig_sota_msra} show that proposed REN obtains the best accuracy among all the algorithms for hand pose estimation.

\begin{figure*}[htb]
\centering
{\includegraphics[width=0.95\textwidth]{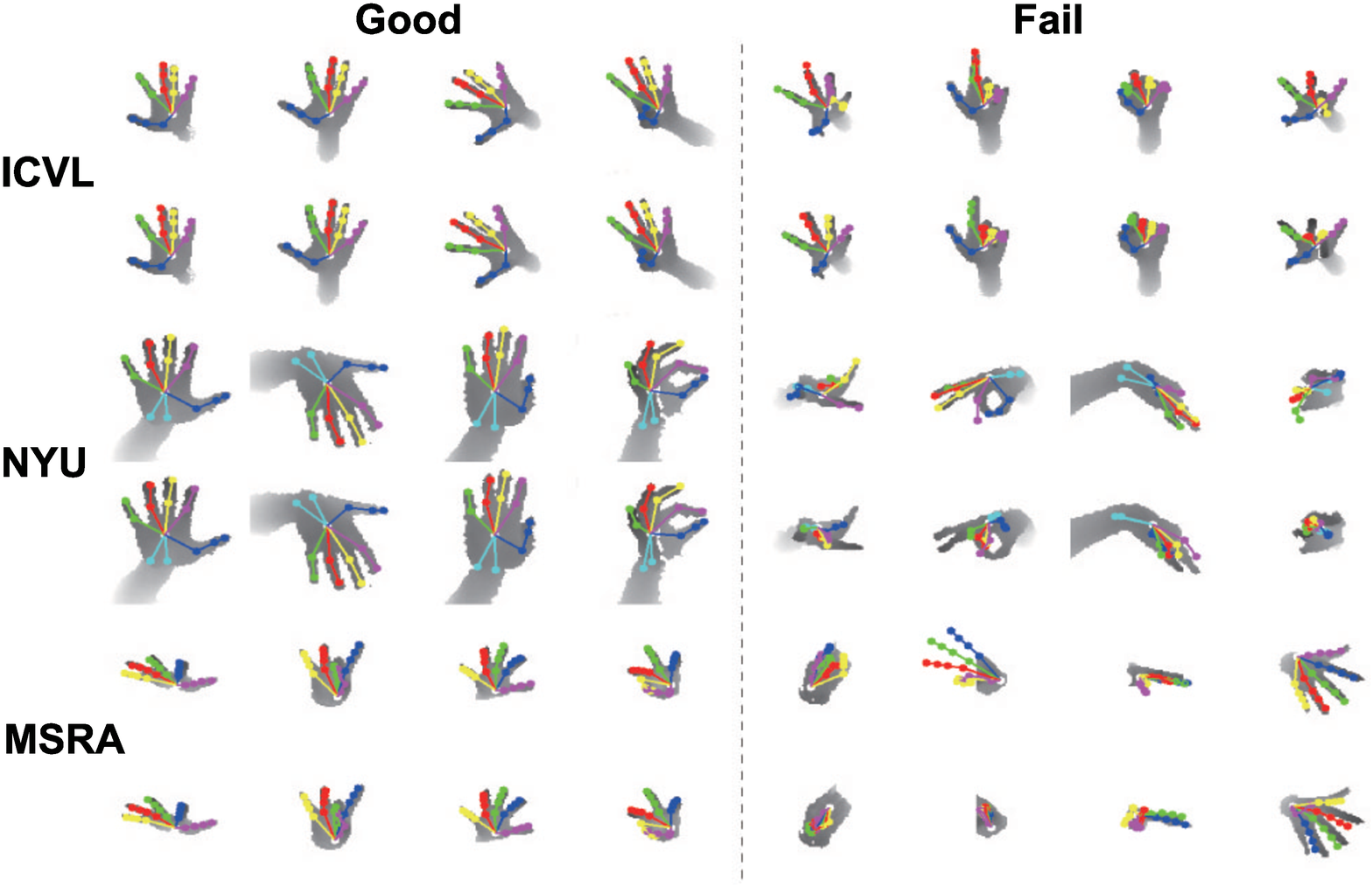}}
\caption{Example results on ICVL \cite{tang2014latent}, NYU \cite{tompson2014real} and MSRA \cite{sun2015cascaded} datasets: ground truth (first row) and region ensemble network (second row) for each datasets.}
\label{fig_vis_all}
\end{figure*}

In details, on ICVL dataset our method surpasses other methods with a large margin. And the mean error $7.31mm$ obtains a $0.80mm$ decrease compared with LSN \cite{wan2016hand}, which is a $9.87\%$ relative improvement. Similarly on NYU dataset, our results are more accurate ($12.69mm$) than other approaches, and reduce the error of \cite{zhang2016learning} by $10.3\%$. For MSRA dataset, our algorithm also significantly outperform all state-of-the-art methods for nearly all thresholds, with an average error of $9.79mm$. Surprisely, it reduces the mean error of \cite{gerobust} by $25.7\%$. Note that either LSN or multi-view ConvNets \cite{gerobust} employ multiple models with complicated design, while our REN only uses single model without multi-stage regression, which indicates the power for proposed region ensemble strategy. Fig. \ref{fig_vis_all} shows some good cases and bad cases for all datasets. We can find that the failure cases are often caused by severe occlusion and bad depth images.

\begin{figure*}[htb]
\begin{minipage}[b]{0.49\textwidth}
  \centering
  \centerline{\includegraphics[width=0.95\textwidth]{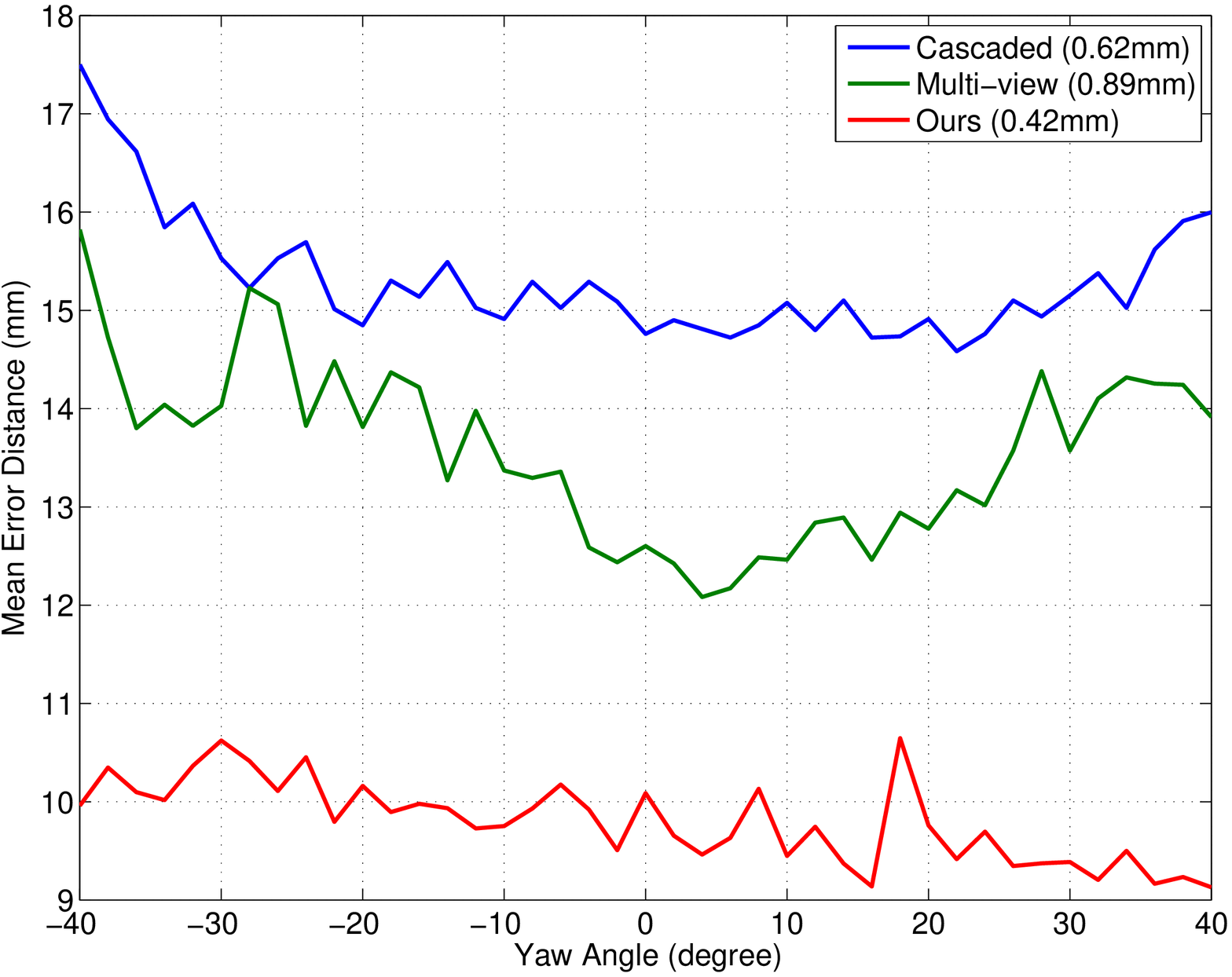}}
\end{minipage}
\begin{minipage}[b]{0.49\textwidth}
  \centering
  \centerline{\includegraphics[width=0.95\textwidth]{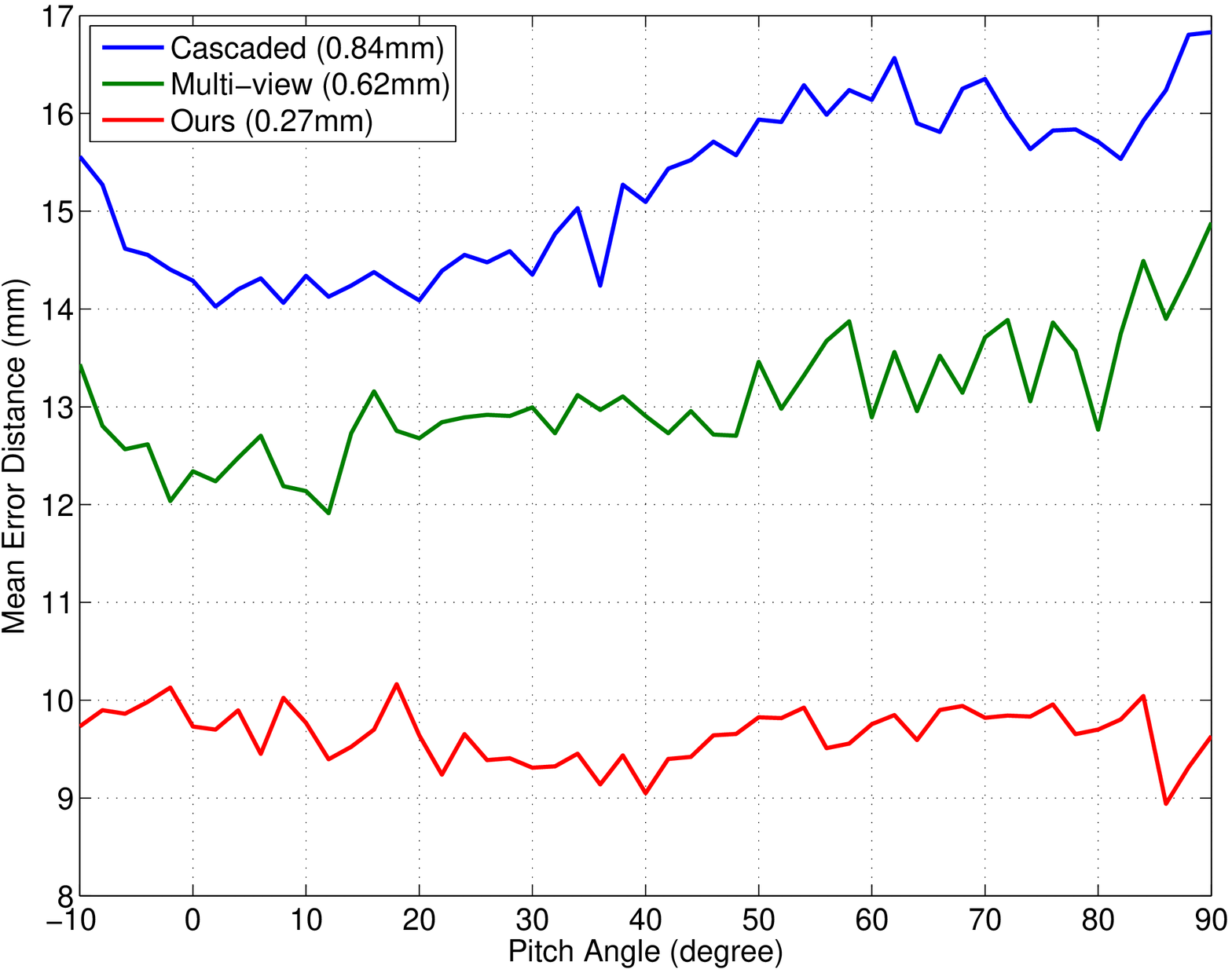}}
\end{minipage}
\caption{The average joint errors distributed over all yaw/pitch viewpoint angles on MSRA \cite{sun2015cascaded} dataset. The standard deviations of the error distributions are shown in the legend titles.}
\label{fig_sota_msra_angle}
\end{figure*}

For MSRA dataset we also report the average joint errors distributed over all yaw and pitch viewpoint angles as in \cite{sun2015cascaded} and \cite{gerobust}, shown in Fig. \ref{fig_sota_msra_angle}. On all angles our method achieve the best accuracy with the smallest deviation, which indicates the robustness for viewpoint variance.

\subsection{Self-comparison}
We perform self comparison experiments for different strategies and setting of region ensemble network on ICVL dataset \cite{tang2014latent}.

\subsubsection{Exploration study}
\begin{figure}[htb]
\centering
{\includegraphics[width=0.48\textwidth]{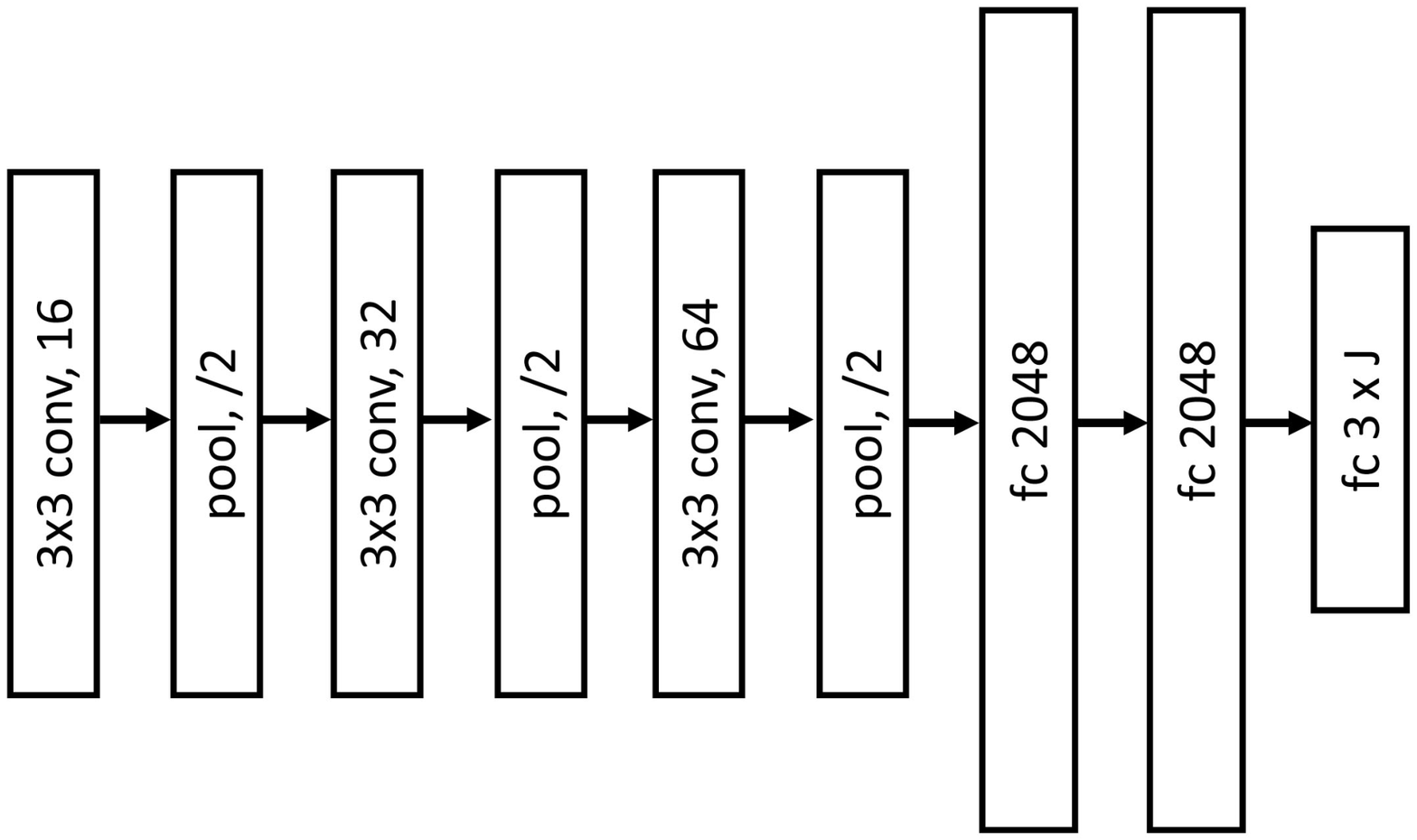}}
\caption{Structure of basic shallow ConvNet with three convolution layers and three pooling layers. The non-linear activation layers following each convolution layers are not showed in the figure.}
\label{fig_shallow}
\end{figure}

In this section, we focus on the investigation of good practices. Specifically, we incrementally introduce five strategies on a basic shallow network in Fig. \ref{fig_shallow}: 1) adding one convolution layer after each convolution layer to increase the depth of ConvNet. 2) adding residual connection edges across pooling layers as described in Section \ref{residual}. 3) using smooth $L_1$ loss \cite{girshick2015fast} instead of Euclidean $L_2$ loss for regression optimization. 4) augmenting the input patches as described in Section \ref{detail}. 5) proposed region ensemble.

The experimental results are summarized in Table 1. Combining all the strategies reduces the errors by 3.17mm (relative $30.2\%$), which is a significant improvement of accuracy. Among them, $L_1$ loss and region ensemble are two most important factors for performance boosting, because $L_1$ loss is more suitable for labels with relative large noise and region ensemble can help improve the generalization for model.

\begin{figure*}[htb]
\centering
{\includegraphics[width=0.95\textwidth]{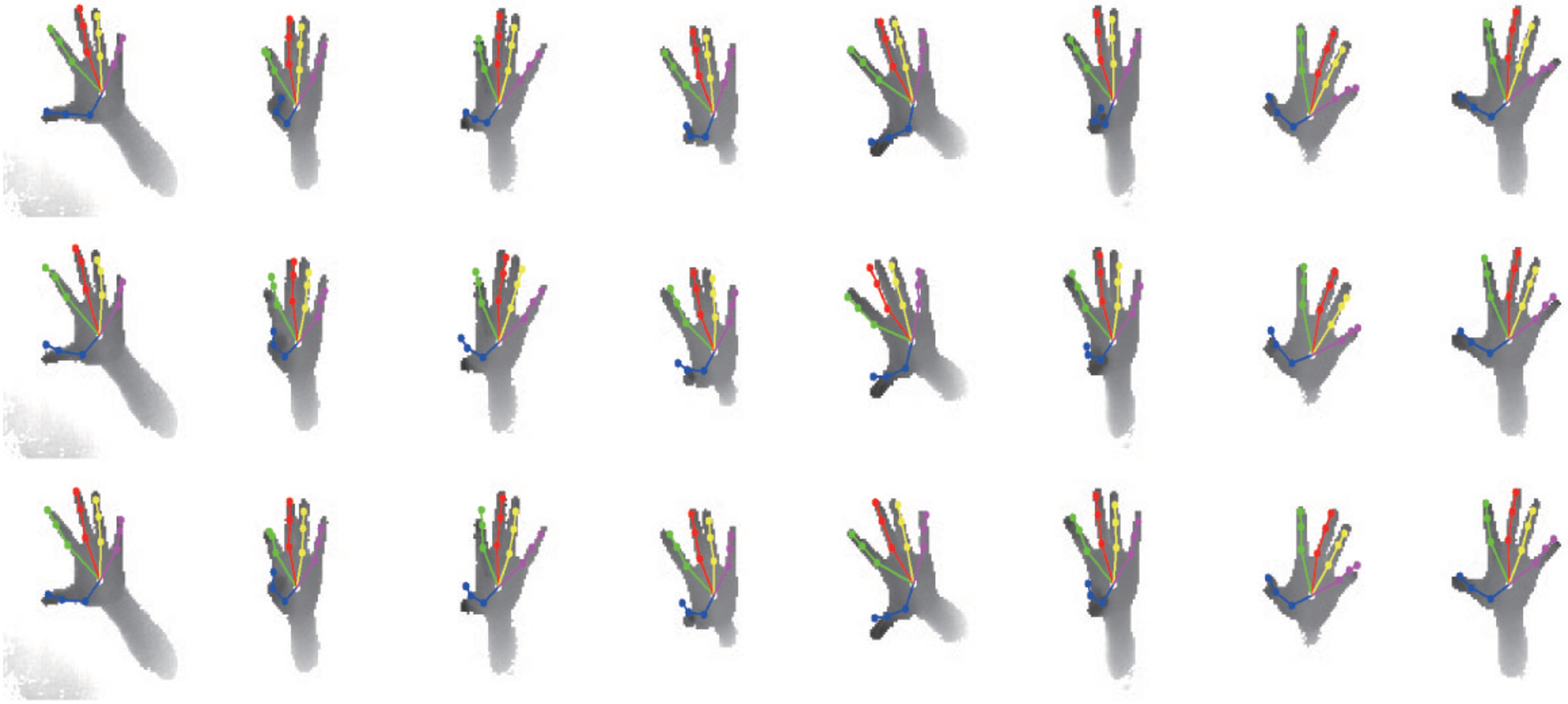}}
\caption{Example results on ICVL \cite{tang2014latent} dataset: ground truth (first row), basic network (second row, corresponding to the fifth row in Table 1), and region ensemble network (third row, corresponding to the seventh row in Table 1).}
\label{fig_vis}
\end{figure*}

Qualitative comparison on ICVL dataset are shown in Fig.\ref{fig_vis} for region ensemble (second row, corresponding to the sixth row in Table 1) and basic network (third row, corresponding to the fifth row in Table 1). The estimations are more accurate for region ensemble especially for fingers.

\begin{table}[htb]
\label{table_add}
\caption{Average 3D distance error (mm) of incremental strategies on ICVL dataset \cite{tang2014latent}. Lower is better.}
\centering
\begin{tabular}{|l|c|}
\hline
\multicolumn{1}{|c|}{Strategy} & Error(mm) \\
\hline
\multicolumn{1}{|c|}{Shallow} & 10.48 \\
+Deeper & 10.02 \\
+Residual Edge & 9.73 \\
+Smooth $L_1$ Loss & 8.59 \\
+Augmentation & 8.36 \\
+Region Ensemble & \textbf{7.31} \\
\hline
\end{tabular}
\end{table}

\subsubsection{Region setting}
\label{setting}
According to the analysis in Section \ref{ensemble}, different region partitions are equal to different patterns of multi-view inputs. Here we explore the effect of different settings of regions, including: 1) multi-scale regions with three regions of size $12\times12$, $8\times8$ and $4\times4$, which is similar to multi-scale inputs as in \cite{tompson2014real} \cite{oberweger2015hands}. 2) four regions of size $6\times6$ (left parts in Fig. \ref{fig_region}), which is the setting in \cite{guo2017region}. 3) nine regions of size $6\times6$ (four as left parts, four as middle parts in Fig. \ref{fig_region} and one in the center), which is the setting in this paper. 4) nine regions of size $4\times4$ with similar positions as (3). 5) nine regions of size $8\times8$ with similar positions as (3).

\begin{figure}[htb]
\centering
{\includegraphics[width=0.48\textwidth]{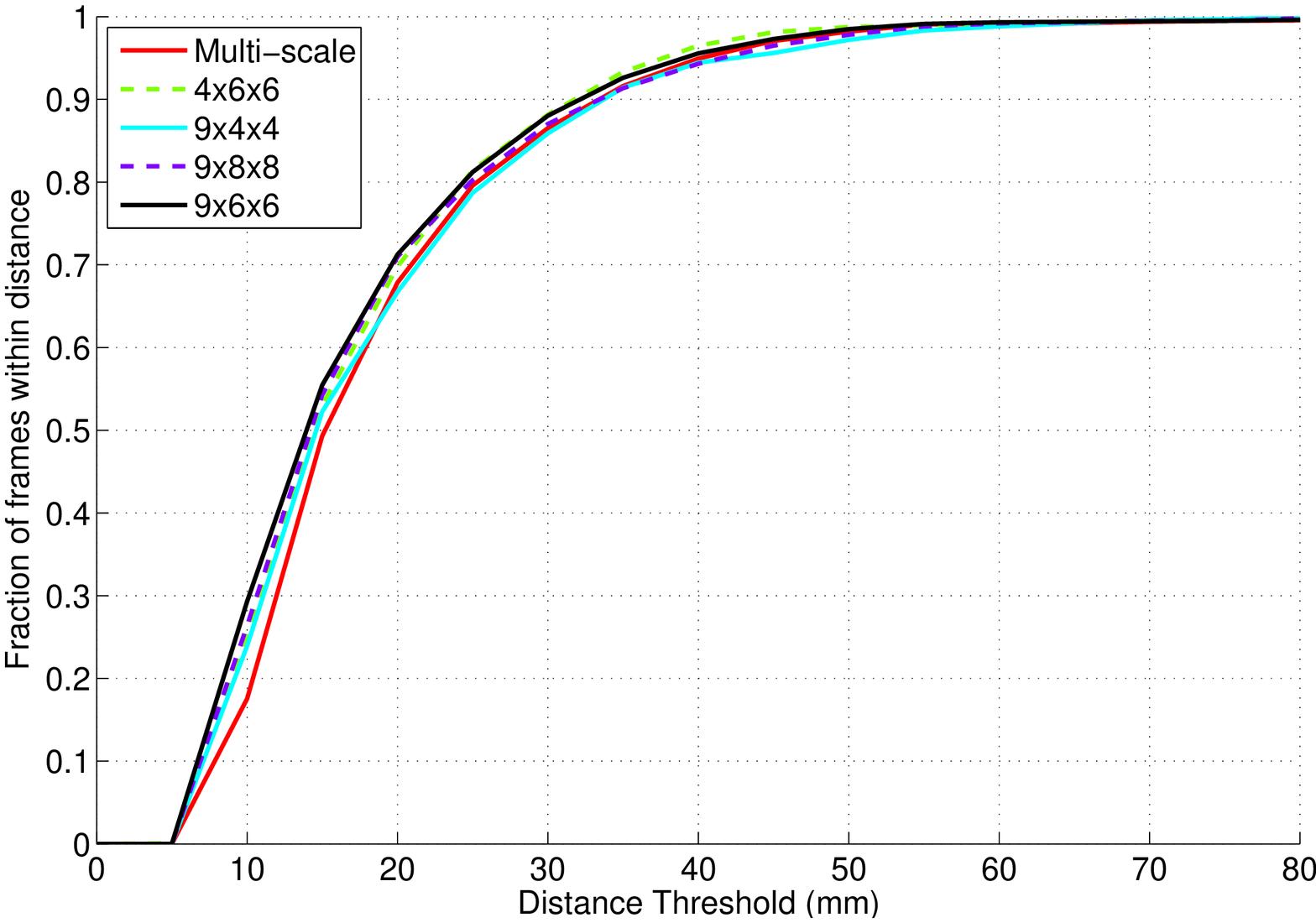}}
\caption{Comparison of different region settings (region number $\times$ region width $\times$ region height) for percentage of success frames on ICVL dataset \cite{tang2014latent}. }
\label{fig_grid}
\end{figure}

From Fig. \ref{fig_grid}, regions with same size are significantly more accurate than multi-scale regions due to the balance parameter number for FC layers of different regions. And more regions with moderate size (i.e. $9\times6\times6$) obtain slightly better performance. Too large or too small receptive field hurts the accuracy of hand pose estimation.

\subsubsection{Comparison with ensembles and multi-view testing}
\label{basic}
We compare with traditional ensembles and multi-view testing in this section. In details, we implement three baselines: 1) \emph{Basic} network has the same convolution structure in Fig.\ref{fig_baseline} and single regressor on the full feature map with two FC layers of 2048 dimensions. 2) \emph{Basic Bagging} network has nine basic networks as (1) that trained independently on the same data with different random order and augmentation. The average predictions of all the networks form the final prediction. 3) \emph{Multi-view Testing} trains single basic network as (1) but averages the predicted 3D hand poses with nine multi-view inputs. The inputs are cropped as in Section \ref{detail}, but on different centers with bias of $-d$/$0$/$d$mm on their x and y coordinates relative to the centroid. We use $d=26.5625$ to approximately match the nine region positions in REN.

\begin{figure}[htb]
\centering
{\includegraphics[width=0.48\textwidth]{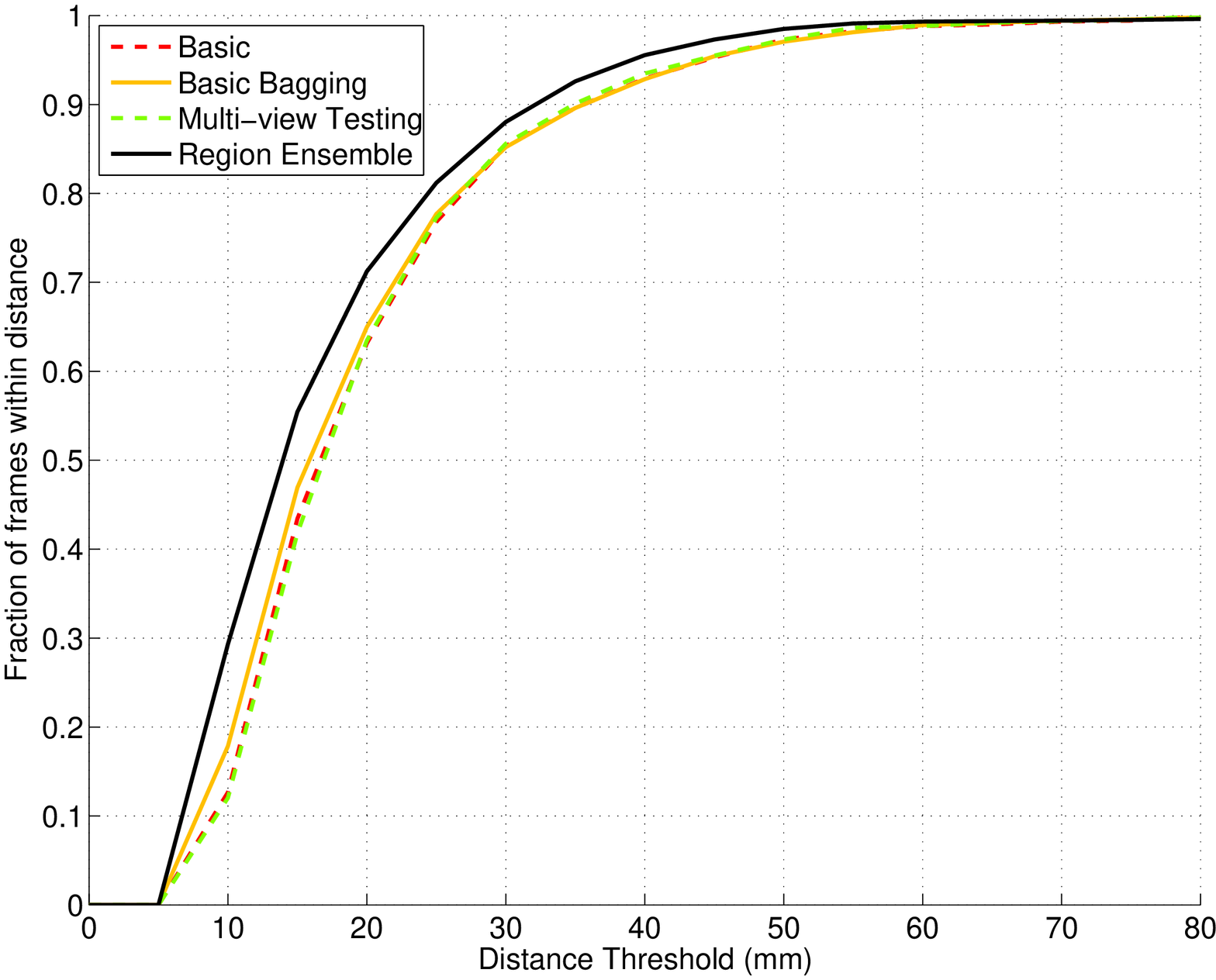}}
\caption{Comparison of ensembles and mutli-view testing for percentage of success frames on ICVL dataset \cite{tang2014latent}. }
\label{fig_ensemble}
\end{figure}

Results in Fig.\ref{fig_ensemble} shows that ensemble based methods (both basic bagging and region ensemble) are significantly more effective that baseline network. And the performance of our region ensemble is much better than traditional bagging. Because REN only employs multiple FC layers instead of multiple complete ConvNets, it also costs less time and memory than traditional bagging. Meanwhile, the improvement from multi-view testing is limited for hand pose estimation, because the model is more sensitive to translation in regression tasks than that in classification.

%
%

\subsection{Evaluation on other RGB-D tasks}
Here we also test our REN on challenging benchmarks for fingertip detection and human pose estimation and compare with state-of-the-art methods.
\subsubsection{Fingertip detection}
We compare the fingertip detection results to several state-of-the-art algorithms \cite{wetzler2015rule} \cite{guo2016two} on NYU dataset without retraining our REN model. Table 3 illustrates that our REN achieves the best performance among all the methods, with an average error of 15.6mm.

\begin{table}[htb]
\label{table_finger}
\caption{Mean precision (mP) and average 3D distance error (mm) for fingertips of different methods on NYU dataset \cite{tompson2014real}. Higher is better for mP and lower is better for error.}
\centering
\begin{tabular}{|c|c|c|}
\hline
Methods & mP & Error(mm) \\
\hline
CNN-DeROT \cite{wetzler2015rule} & 0.63 & - \\
DeepPrior \cite{oberweger2015hands} & 0.43 & 26.4 \\
FeedLoop \cite{oberwegertraining} & 0.38 & 23.2 \\
TwoStream \cite{guo2016two} & 0.50 & 19.3 \\
Model \cite{zhou2016model} & 0.40 & 24.4 \\
REN & \textbf{0.66} & \textbf{15.6} \\
\hline
\end{tabular}
\end{table}

\subsubsection{Human pose estimation}
\begin{figure*}[htb]
\centering
{\includegraphics[width=0.95\textwidth]{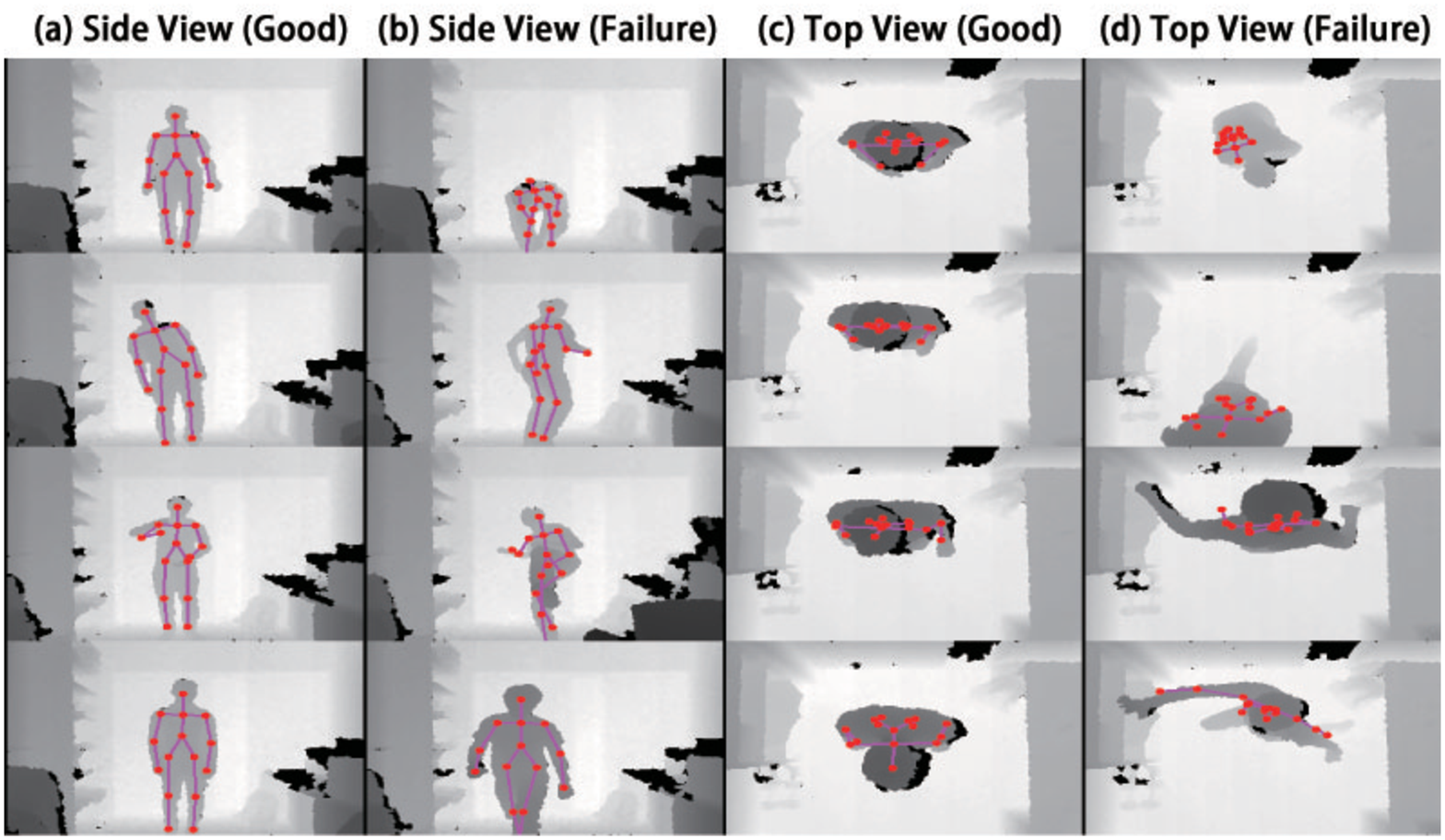}}
\caption{Example results on ITOP \cite{carreira2016human} dataset.}
\label{fig_itop_vis}
\end{figure*}

The results for human pose estimation are reported in Table 4, where we compare our method with \cite{yub2015random} \cite{carreira2016human} using mAP metric on ITOP dataet. For frontal view, proposed REN with 84.9 mAP significantly outperforms RTW and REF. And the accuracy for lower body is much higher. For top-down view, our method is better than RTW and shows comparable performance with REF, which contains deeper ConvNets with 16 convolution layers in their models. See Fig. \ref{fig_itop_vis} for some visualization results.
\begin{table*}[htb]
\label{table_human}
\caption{Mean average precision (mAP, unit: \%) of different methods on frontal view and top view of ITOP dataset \cite{haque2016towards} using a 10cm threshold. Higher is better.}
\centering
\begin{tabular}{|c|c|c|c|c|c|c|}
\hline
& \multicolumn{3}{|c|}{mAP (front-view)} & \multicolumn{3}{|c|}{mAP (top-view)} \\
\hline
Body Part & RTW \cite{yub2015random} & REF \cite{carreira2016human} & REN & RTW \cite{yub2015random} & REF \cite{carreira2016human} & REN \\
\hline
Head & 97.8 & 98.1 & 98.7 & 98.3 & 98.1 & 98.2 \\
Neck & 95.8 & 97.5 & 99.4 & 82.2 & 97.6 & 98.9 \\
Shoulders & 94.1 & 96.6 & 96.1 & 91.8 & 96.1 & 96.6 \\
Elbows & 77.9 & 73.3 & 74.7 & 80.1 & 86.2 & 74.4 \\
Hands & 70.5 & 68.6 & 55.2 & 76.9 & 85.5 & 50.7 \\
Torso & 93.8 & 85.6 & 98.7 & 68.1 & 72.9 & 98.1 \\
Hips & 80.3 & 72.0 & 91.8 & 55.7 & 61.1 & 85.5 \\
Knees & 68.8 & 69.0 & 89.0 & 53.9 & 51.6 & 70.0 \\
Feet & 68.4 & 60.8 & 81.1 & 28.6 & 51.5 & 41.6 \\
\hline
Mean & 80.5 & 77.2 & \textbf{84.9} & 68.5 & \textbf{75.5} & \textbf{75.5} \\
\hline
\end{tabular}
\end{table*}

\noindent\textbf{Implementation details}\hspace{2mm} For human pose, a small ConvNet is trained to predict the torso position as center. The size of cube is $800\times1200\times800\textrm{mm}^3$ for front-view and $600\times600\times1000\textrm{mm}^3$ for top-view. For data augmentation, random flip of image with probability of 0.5 is also used. 

\section{Conclusion}
To boost the performance of single ConvNet for 3D hand pose estimation, we exploit several good practices and present a simple but powerful region ensemble structure by dividing the feature maps into different regions and jointly training multiple regressors on all regions with fusion. Such strategies significantly improve the accuracy of ConvNet. The experimental results demonstrate that our method outperforms all the state-of-the-arts on three hand pose datasets and one human pose dataset. Since region ensemble is easy to be introduced into ConvNets, we believe that proposed structure could be applied on more computer vision tasks and achieve more promising results.

\noindent\textbf{Acknowledgments}\hspace{2mm} This work is supported by The National Science Foundation of China (No. 61271390 and No. 91648116), and State High-Tech Development Plan (No. 2015AA016304). Thanks Shulan Pan for paper edition.

{\small
\bibliographystyle{ieee}
\bibliography{refs}
}

\end{document}